\documentclass{article}

\usepackage{arxiv}

\usepackage[utf8]{inputenc}
\usepackage[T1]{fontenc} 
\usepackage{hyperref}    
\usepackage{url}          
\usepackage{booktabs}       
\usepackage{amsfonts}  
\usepackage{nicefrac}    
\usepackage{microtype}
\usepackage{graphicx}
\usepackage{natbib}
\usepackage{doi}
\usepackage{graphicx}
\usepackage{subcaption}
\usepackage{amsmath}
\usepackage{cleveref}

\title{Exploring Clustering Capability of Inpainting Model Embeddings for Pattern-based Individual Identification}

\date{May 6, 2026}

\usepackage{authblk}

\newbox{\orcid}\sbox{\orcid}{\includegraphics[scale=0.06]{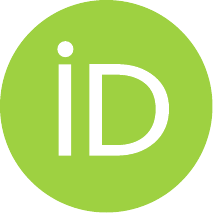}} 
\author[1,2]{%
	\href{https://orcid.org/0009-0006-1705-5595}{\usebox{\orcid}\hspace{1mm}Jens~van~Bijsterveld\thanks{\texttt{j.van.bijsterveld@liacs.leidenuniv.nl}}}%
}
\author[3]{%
	\href{https://orcid.org/0000-0003-3714-7973}{\usebox{\orcid}\hspace{1mm}Daniele Avitabile\thanks{\texttt{d.avitabile@vu.nl}}}%
}
\author[1]{%
	\href{https://orcid.org/0000-0003-2445-8158}{\usebox{\orcid}\hspace{1mm}Fons Verbeek\thanks{\texttt{f.j.verbeek@liacs.leidenuniv.nl}}}%
}
\author[1,2]{%
	\href{https://orcid.org/0000-0002-2970-1180}{\usebox{\orcid}\hspace{1mm}Rita Pucci\thanks{\texttt{r.pucci@liacs.leidenuniv.nl}}}%
}
\affil[1]{Leiden Institute of Advanced Computer Science, Leiden University, Leiden, The Netherlands}
\affil[2]{Naturalis Biodiversity Center, Leiden, The Netherlands}
\affil[3]{Department of Mathematics, Vrije Universiteit, Amsterdam, The Netherlands}

\renewcommand{\shorttitle}{Inpainting Model Embeddings for Pattern-based Individual Identification}

\hypersetup{
pdftitle={Exploring Clustering Capability of Inpainting Model Embeddings for Pattern-based Individual Identification},
pdfauthor={Jens~van~Bijsterveld, Daniele~Avitabile, Fons~Verbeek, Rita~Pucci},
pdfkeywords={Representation Learning, Inpainting, Individual Identification, Clustering, Skin Patterns},
}

\begin{document}
\maketitle

\begin{abstract}
  In this paper, we explore deep learning techniques for individual identification of animals based on their skin patterns. Individual identification is crucial in biodiversity monitoring, since it enables analysis of decline or growth of populations, or intra-species interactions within populations.
  Models trained for the task of individual identification often do not focus on the skin pattern of animals, but on background details or body shape details. These characteristics are not individually specific, or can change drastically through time. We focus on techniques that will make machine learning models more responsive to skin pattern structure when extracting individual visual embeddings from images. For this, we explore image inpainting of task-specific masks as an auxiliary task to enhance ML-based individual identification from animal skin patterns.
  We propose a comparative analysis among four models as an encoder backbone for the individual identification task. We focus on the case study of zebrafish, which is a widely recognized biological model organism, and which exhibits individually identifying skin patterns. 
  To evaluate encoder backbone performance, we present standard metrics for classification accuracy, embedding clustering metrics, and GradCAM visualizations.
\end{abstract}

\keywords{Representation Learning \and Inpainting \and Individual Identification \and Clustering \and Skin Patterns}

\section{Introduction} \label{sec:intro}

The Earth is facing an unprecedented decline in biodiversity~\cite{diaz_pervasive_2019}. In order to mitigate this biodiversity crisis, conservation efforts are being ramped up. Global commitments to halt the biodiversity decline have been made, e.g. the Aichi Biodiversity Targets~\cite{unit_aichi_2020} and the United Nations Sustainable Development goals~\cite{UN_Sustainable_Development_17_nodate}. To keep up with the required conservation effort for these goals, more and more technologically advanced techniques are needed in the field of biodiversity monitoring~\cite{freitas_biodiversity_2025}. Our focus will be on techniques for speeding up the labelling and analysis of digital data (e.g. images).

In fact, many advanced data collection techniques are already being deployed in the field. Camera trapping is used for collecting visual data on animal presence, behaviour, and distribution~\cite{rovero_camera_2016}, drones are being deployed to survey, map, and monitor biodiversity and habitat conditions cheaply~\cite{wich_conservation_2018}, and acoustic monitoring is used for long-term, cost-effective monitoring~\cite{sugai_terrestrial_2019}. There are also contributions in the field of environmental DNA collection~\cite{deiner_environmental_2017} and citizen science~\cite{chandler_contribution_2017,molls_obs-services_2021}. As such, there is a large increase in available biodiversity conservation data, which is infeasible to process manually.

Computer Vision provides a way to automatically process and interpret large amounts of visual data by, for example, identifying objects from the background~\cite{khanam_yolov11_2024}, tracking objects through a field of view~\cite{kadam_object_2024}, or classifying object identities~\cite{li_analysis_2023}. The classification of animals from image data currently performs quite well on the species level~\cite{gomez_villa_towards_2017,ahmed_animal_2020, waldchen_plant_2018, waldchen_machine_2018}, but classification performance on within-species individual identification (outside controlled conditions) is still an open challenge~\cite{schneider_similarity_2022}.

Previously, individual animals were tracked through invasive measures, such as placing a tag, or collecting DNA~\cite{krebs_ecological_1989}. Visual identification based on animal biometrics, e.g. iris pattern, facial features, retinal vascular pattern, gait, or body patterns, would improve upon this, since it is a non-invasive, less labour-intensive method of tracking~\cite{kuhl_animal_2013, allen_assessing_2015}. Visual identification is, however, still challenging, since most animal biometrics are hard to accurately capture, prone to changes during an animal's lifespan, or reliant on domain experts~\cite{cihan_identification_2023}.

One of the more easily available and informative animal biometrics is skin patterning. Animals like, for example, leopards, jaguars~\cite{liu_two-stage_2006}, ladybugs~\cite{liaw_turing_2001}, and zebrafish~\cite{bullara_pigment_2015} all exhibit distinctive individual patterns. These patterns all emerge through autonomous interactions between groups of cells, and have been studied extensively by developmental biologists~\cite{reeves_quantitative_2006}.

In this analysis, we explore a method for embedding information on biologically significant patterns in encoders, with the final application of individual identification from raw image data. For this, we explore image inpainting as an auxiliary task to enhance individual identification using machine learning based on animal skin patterns. We analyse classification accuracy, model focus when extracting features, and embedding space clustering on a dataset of zebrafish images.

\section{Related Work}

In the past, individual (re-)identification in biodiversity monitoring used to be done through tagging. This process is invasive, labour intensive, and expensive, since they require catching the animal in question, and physically tagging or identifying them~\cite{krebs_ecological_1989}. Non-invasive individual identification of animals, through e.g. visual identification based on biometrics, is therefore better and more efficient~\cite{vidal_perspectives_2021}.

Manual visual identification of animals is error-prone, since it requires domain experts, and identification is biased based on the observer. Therefore, in previous work, images were mathematically analysed, and used for the creation of handcrafted features~\cite{hiby_tiger_2009,allen_assessing_2015}. These features, for example tomographic characteristics of stripes or speck distance on skin, were then used to individually identify animals~\cite{cisar_computer_2021,pedersen_finding_2023}. However, this method is species-specific, and not generalizable to different species.

A more generalizable method of feature extraction can be achieved through the application of deep learning~\cite{simonyan_very_2015}. Convolutional neural networks and vision transformers use learned weights in order to extract features from images. These learned features do, however, not necessarily have to be informative for the animals that are to be studied. For example, models trained for the task of individual identification often do not focus on the skin pattern of the animal but on background details or body shape details of the animals~\cite{nepovinnykh_species-agnostic_2024}. These characteristics are not individual specifics, or can change drastically through time.

There is, therefore, a gap in the literature for teaching models to use informative features for the individual identification of animals. In previous work, Puchalla, Serianni, and Deng~\cite{puchalla_zebrafish_2025} performed an ablation study on a Convolutional Neural Network (CNN) and Vision Transformer (ViT) for individual zebrafish classification, showing that features located on the body of the fish were the most influential, but no specific inquiry with regard to the skin patterns was made. Skin patterns are a very promising feature, since they are, in many animal species, truly unique for each individual.

Zebrafish (\textit{Danio rerio}) is a widely used model organism in molecular biology and developmental biology, and therefore widely used in skin pattern development studies~\cite{kondo_studies_2021}. This wide usage of zebrafish contributes to the availability of suitable datasets.

In previous work, individual zebrafish were mostly identified based on motion tracking. This was either performed manually~\cite{mcelligott_prey_2005}, based on the regression-head of an object detector network~\cite{gudiksen_regression_2021}, or based on the fish direction through automated detection of zebrafish larvae heads~\cite{si_tracking_2022}. Static individual identification of zebrafish has so far been done through the use of pre-computed features such as Keep It Simple and Straightforward Metric (KISSME)~\cite{kostinger_large_2012}, improved KISSME (iKISSME)~\cite{yang_large_2016}, Large Scale Similarity Learning (LSSL)~\cite{yang_large_2016}, Cross-view Quadratic Discriminant Analysis (XQDA)~\cite{liao_person_2015}, and Kernelized Discriminative Null Space (DNS)~\cite{zhang_learning_2016}~\cite{haurum_re-identification_2020}, and CNN and ViT classification based on a rolling window training set~\cite{puchalla_zebrafish_2025}. More recently, a dual-stage model based on InceptionV3~\cite{szegedy_rethinking_2015} and a 32 by 32 vision transformer~\cite{dosovitskiy_image_2021} was used to successfully identify zebrafish during their development. Skin pattern was shown to have a large contribution to successful classification~\cite{cao_longitudinal_nodate}. 

Our work will go beyond the current state of the art by exploring the possibility of pattern-based individual identification without images from different developmental stages. Classification will be done purely based on inferred pattern characteristics, allowing future generalization to different specimens and different species.

Generative Adversarial Networks (GANs) try to capture the distribution of the training data in order to be able to generate samples that are as close to it as possible~\cite{goodfellow_generative_2014}. This causes them to postulate hidden variables as the causes of the complicated dependencies in visual patterns. Adversarially trained Masked Auto Encoders (MAEs) are a special case of Variational Auto Encoders (VAEs)~\cite{kingma_auto-encoding_2022} that learn efficient and meaningful representations of data by selectively masking and reconstructing input features~\cite{he_masked_2022}.

MAEs are a good choice for capturing dependencies between image features if there is a relatively small amount of data available~\cite{pathak_context_2016}. This property has been used in previous work for pre-training for bird species classification and localization~\cite{sastry_birdsat_2024}, and in order to create a forest plant classification network~\cite{huy_masked_2024}.

In this work we compare four different state-of-the-art inpainting models on their ability to capture pattern information for the task of pattern-based individual identification of zebrafish. All selected models are encoder-decoder and GAN based, use purely convolutional encoders, and emphasize context modelling and receptive field.

AOT-GAN~\cite{zeng_aggregated_2021} uses Aggregated Contextual Transformation (AOT) blocks, which aggregate contextual information across multiple receptive fields, allowing the model to capture both distant global context and local image patterns necessary for realistic reconstruction. It also uses an improved version of the adversarial network. AOT-GAN was chosen since its multiscale convolutional receptive fields could aggregate multiple important pattern features.

DeepFillV2~\cite{yu_free-form_2019} uses gated convolutions, which introduce learnable dynamic feature selection mechanisms that compute gating values for each channel and spatial location, allowing the network to selectively propagate valid and relevant features. This architecture was chosen, since selective propagation might allow DeepFillV2 to specifically select the most important parts of provided patterns.

EdgeConnect~\cite{nazeri_edgeconnect_2019} uses a two-stage network for inpainting. The first network, the edge generator, infers and completes edges from the masked image, which the second network, the inpainting generator, uses to accurately predict the contents of the mask. EdgeConnect was chosen since its focus on edges should make it very compatible with patterns.

LaMa~\cite{suvorov_resolution-robust_2022} uses fast Fourier convolutions, a specific type of spectral residual learning block~\cite{chi_fast_2019}, in order to enlarge its receptive field. This large receptive field, together with LaMa's proven high performance on reconstructing regions with challenging geometric patterns, are expected to produce informative pattern embeddings.

Furthermore, all chosen inpainting models outperformed all previously proposed inpainting models with regard to Peak Signal-to-Noise Ratio, Structural SIMilarity index (SSIM)~\cite{wang_image_2004}, and mean absolute error, also called L1 loss (and Fréchet Inception Distance~\cite{heusel_gans_2017} for EdgeConnect, AOT-GAN, and LaMa) at time of publishing on the places2 benchmark dataset~\cite{zhou_places_2018}, which is a dataset containing images of places like buildings, which in turn include repetitive pattern-like structures. 

Inter-model comparisons are provided for AOT-GAN, which outperforms DeepFillV2 and EdgeConnect~\cite{zeng_aggregated_2021}, and LaMa, which outperforms all three other models~\cite{suvorov_resolution-robust_2022}.
Pre-trained checkpoints on the places2 dataset are provided for each of the selected models.

\section{Methodology} \label{sec:methodology}

\begin{figure}[tb]
  \centering
  \includegraphics[height=5cm]{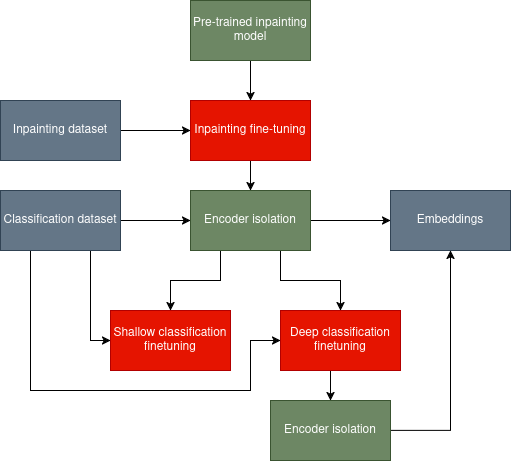}
  \caption{Graphical overview of the proposed experiments. Architectures are pre-trained on the inpainting task, and then their encoders are fitted with a classification head, and fully or partially fine-tuned for individual classification.}
  \label{fig:method-overview}
\end{figure}

This section describes the methodology for the experiments constituting this exploratory analysis. An outline of the proposed experiments can be seen in \cref{fig:method-overview}.

\subsection{Dataset collection} \label{sec:dataset-collection}

Zebrafish image data was collected from the work by Haurum, Karpova, Pedersen \emph{et al}.~\cite{haurum_re-identification_2020}.

For the inpainting dataset, binary masks occluding a large part of the skin patterns on the zebrafish bodies were manually created. Different images were selected for the classification dataset, and paired with empty masks for compatibility.

\subsection{Inpainting Task} \label{sec:inpainting-method}

During the inpainting task, the four selected inpainting models were used to capture dependencies between image features within their respective architectures.

The inpainting task is defined as the creation of artificial image $\hat{I} \in \mathbb{R}^{H \times W \times C}$ from the original image $I \in \mathbb{R}^{H \times W \times C}$, where some regions are corrupted by multiplication with a binary mask $M \in \mathbb{R}^{H \times W}$, creating input image $X = I \times M^{-1}$. In this paper, the reconstruction of $I$ through the creation of $\hat{I}$ is done by passing the tensor $X \oplus M$ through the inpainting model $f$ with learned parameters $\phi$. The full equation for the inpainting task can be seen in \cref{eq:inpainting-task}. Model weights were initialized from prior training, and updated through backpropagation based on the loss functions proposed by the models' respective authors.

\begin{equation}
  \hat{I} = f_\phi((I \times M^{-1}) \oplus M)
  \label{eq:inpainting-task}
\end{equation}

In preparation of inpainting training, the AOT-GAN~\cite{noauthor_researchmmaot-gan-for-inpainting_2026}, EdgeConnect~\cite{nazeri_knazeriedge-connect_2026}, and LaMa (big-LaMa version)~\cite{noauthor_advimmanlama_2026} models were cloned from their respective repositories, while DeepFillV2 was cloned from a PyTorch fork of the original repository~\cite{nipponjo_nipponjodeepfillv2-pytorch_2026}. EdgeConnect and DeepFillV2 code were modified to use predetermined deterministic masks (manually created during dataset collection) instead of their respective included random mask generators. These alterations were not needed for AOT-GAN and LaMa, which already included the option to load in pre-defined masks. Pre-trained model checkpoints for the places2 dataset~\cite{zhou_places_2018} were loaded into the models.

All models were fine-tuned for inpainting zebrafish skin patterns for 10 epochs with a batch size of 8. For encoder isolation and subsequent classification fine-tuning, the checkpoint with the lowest loss on the validation dataset during training was used (\cref{tab:inpainting-checkpoints}). Used loss functions were the common L1 reconstruction loss, the (Hinge) Adversarial loss~\cite{goodfellow_generative_2014}~\cite{lim_geometric_2017}, Perceptual loss~\cite{johnson_perceptual_2016}, and (for EdgeConnect) Feature matching loss~\cite{sajjadi_enhancenet_2017}. For each of the inpainting models, the generator loss was calculated as a weighted sum of these loss functions~\cite{zeng_aggregated_2021,yu_free-form_2019,nazeri_edgeconnect_2019,suvorov_resolution-robust_2022}.

After inpainting training, we isolated the encoder of each model, defined here as $g_\phi$. The AOT-GAN encoder consists of padding, followed by three times a convolutional layer and a nonlinear ReLU activation function. For DeepFillV2, the encoder consists of five gated convolutions, and for EdgeConnect, the encoder consists of a padding operation, followed by three times a convolution and batch normalization, followed by a ReLU. The LaMa encoder consists of a padding operation, followed by three standard convolutional layers with batch normalization and ReLU, and ends with a convolutional block consisting of a global and a local pathway.

Isolated encoders were used to generate embeddings $E = g_\phi(I \oplus \mathbf{0})$ for all images $I$ in the classification test dataset. Note that a full-zero mask $\mathbf{0}$ was concatenated to the input images for compatibility with the isolated encoders.

\subsection{Classification Task} \label{sec:classificationTask}

During the classification task, captured pattern information is coupled to information on individual identity.

Starting from $g_\phi()$, a classification model was created by adding a classification head consisting of a linear layer ($LL$) with the input size defined by the encoder output and the output size defined by the number of output classes ($K$), and a softmax activation function $\sigma : \mathbb{R}^K \rightarrow (0, 1)^K$ for normalization into a probability distribution $p \in \mathbb{R}^K$.

Resulting classification models were trained with either shallow backpropagation, consisting of a frozen encoder with a trainable classification head (\cref{eq:shallow}), or full backpropagation, consisting of a fully trainable encoder and classification head (\cref{eq:deep}). Training was performed for 15 epochs with a batch size of 8 on the classification training set, with performance metrics computed each epoch on the validation set. Model feature extraction focus was also computed each training epoch by using GradCAM~\cite{selvaraju_grad-cam_2020, gildenblat_jacobgilpytorch-grad-cam_2026}. After training, embeddings were generated for the deep backpropagation classification task, using the previously defined split encoder models with classification fine-tuned weights.

\begin{equation}
  p = \sigma(LL_\theta(g_{\text{frozen}}(I \oplus \mathbf{0})))
  \label{eq:shallow}
\end{equation}

\begin{equation}
  p = \sigma(LL_\theta(g_\theta(I \oplus \mathbf{0})))
  \label{eq:deep}
\end{equation}

The loss used during classification fine-tuning is the Cross-Entropy loss. Classification is evaluated using accuracy, recall and F1-score.

\subsection{Embedding clustering}

In order to evaluate the encoder backbones' pattern-based generalization capability, a clustering analysis was performed on the embeddings computed after inpainting and classification training.

We performed a qualitative analysis by reducing the computed embeddings to 2D through PCA~\cite{mackiewicz_principal_1993}, T-SNE~\cite{maaten_visualizing_2008}, and UMAP~\cite{mcinnes_umap_2020}, and a quantitative analysis by clustering embeddings using the k-means algorithm~\cite{lloyd_least_1982}, and computing the Silhouette Score~\cite{rousseeuw_silhouettes_1987}, Davies-Bouldin Index~\cite{davies_cluster_1979}, Calinski-Harabasz Index~\cite{calinski_dendrite_1974}, Adjusted Rand Index~\cite{rand_objective_1971}, and Mutual Information~\cite{shannon_mathematical_1948}.

\section{Ablation Study}

During the ablation study, we evaluate the most important image region for the proposed inpainting pre-trained classification networks.

Encoders were initialized with weights from the inpainting task and fitted with a classification head as described in \cref{sec:classificationTask}. These architectures were then fine-tuned (with both deep and shallow backpropagation) on 300 images per individual zebrafish originating from the classification test set with a manually masked background, fish, or pattern region. EfficientNet~\cite{tan_efficientnet_2020} and InceptionV3~\cite{szegedy_rethinking_2015}, as initialized from their respective PyTorch checkpoints, were also fine-tuned on the masked region data.

\begin{figure}[tb]
  \centering
  \begin{subfigure}[t]{0.32\linewidth}
    \includegraphics[width=\linewidth]{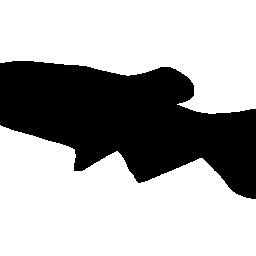}
    \caption{Background area mask}
    \label{fig:bg}
  \end{subfigure}
  \hfill
  \begin{subfigure}[t]{0.32\linewidth}
    \includegraphics[width=\linewidth]{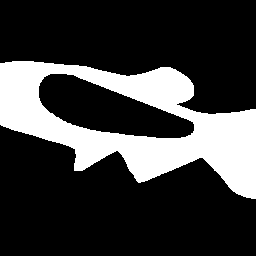}
    \caption{Fish area mask}
    \label{fig:fish}
  \end{subfigure}
  \hfill
  \begin{subfigure}[t]{0.32\linewidth}
    \includegraphics[width=\linewidth]{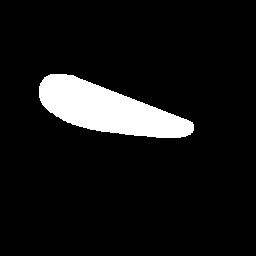}
    \caption{Pattern area mask}
    \label{fig:pattern}
  \end{subfigure}
  \caption{Hand-drawn masks for the background, fish, and pattern areas as used for the ablation study. The background area mask consists of the full image background, the fish area mask consists of the perimeter and some minor body pattern of the fish, and the pattern area mask consists of a substantial part of the body pattern of the fish.}
  \label{fig:ablation-data}
\end{figure}

An example of the masks used in this ablation study can be seen in \cref{fig:ablation-data}. Accuracy metrics for the different ablations are shown in \cref{tab:ablation-deep} and \cref{tab:ablation-shallow} for deep and shallow backpropagation, respectively. The unmasked region(s) leading to the best model performance are highlighted in a \textbf{bold font}. The worst performing area(s) are highlighted in \textit{italics}.

\begin{table}
  \centering
  \caption{Accuracy for each region and combination of regions tested during the ablation study. Fine-tuning was performed using deep backpropagation.}
  \label{tab:ablation-deep}
  \begin{tabular}{l|lll|lll|l}
    \toprule
    Algorithm & Background & Fish & Pattern & No background & No fish & No pattern & All \\
    \midrule
    AOT-GAN & \textit{0.75} & \textbf{0.88} & 0.85 & \textit{0.82} & 0.89 & \textbf{0.93} & 0.97 \\
    DeepFillV2 & 0.68 & \textbf{0.75} & \textit{0.63} & 0.75 & \textbf{0.89} & \textit{0.68} & 0.97 \\
    EdgeConnect & \textit{0.64} & \textbf{0.93} & 0.78 & \textbf{0.97} & 0.82 & \textit{0.60} & 0.99 \\
    LaMa & \textit{0.89} & \textbf{0.96} & 0.93 & \textbf{0.99} & 0.98 & \textit{0.94} & 0.99 \\
    EfNet & \textbf{\textit{0.17}} & \textbf{\textit{0.17}} & \textbf{\textit{0.17}} & \textbf{\textit{0.17}} & \textbf{\textit{0.17}} & \textbf{\textit{0.17}} & \textbf{\textit{0.17}} \\
    Inception & \textbf{\textit{0.17}} & \textbf{\textit{0.17}} & \textbf{\textit{0.17}} & \textbf{\textit{0.17}} & \textbf{\textit{0.17}} & \textbf{\textit{0.17}} & \textbf{\textit{0.17}} \\
    \bottomrule
  \end{tabular}
\end{table}

\begin{table}
  \centering
  \caption{Accuracy for each region and combination of regions tested during the ablation study. Fine-tuning was performed using shallow backpropagation.}
  \label{tab:ablation-shallow}
  \begin{tabular}{l|lll|lll|l}
    \toprule
    Algorithm & Background & Fish & Pattern & No background & No fish & No pattern & All \\
    \midrule
    AOT-GAN & \textbf{0.76} & 0.32 & \textit{0.17} & \textit{0.46} & \textbf{0.86} & 0.68 & 0.75 \\
    DeepFillV2 & 0.67 & \textbf{0.69} & \textit{0.63} & 0.75 & \textbf{0.76} & \textit{0.64} & 0.60 \\
    EdgeConnect & \textit{0.44} & \textbf{0.95} & 0.83 & \textit{0.17} & \textbf{0.82} & 0.76 & 0.97 \\
    LaMa & \textit{0.86} & \textbf{0.97} & 0.94 & \textbf{0.99} & \textit{0.94} & 0.97 & 0.99 \\
    EfNet & \textit{0.76} & \textbf{0.93} & \textbf{0.93} & \textbf{0.97} & \textit{0.92} & \textit{0.92} & 0.96 \\
    Inception & \textit{0.69} & 0.85 & \textbf{0.88} & \textbf{0.94} & \textit{0.79} & 0.82 & 0.89 \\
    \bottomrule
  \end{tabular}
\end{table}

For deep backpropagation classification fine-tuning, we observe that the \texttt{background} region is the least informative for all architectures except DeepFillV2 (for which the \texttt{pattern} region is the least informative). We also see that the fish region is the most informative for all architectures.

For combinations of areas with deep backpropagation, we see that \texttt{no pattern} performs worst for all architectures except AOT-GAN (for which missing the \texttt{background} region performs worst). We also see that the combination of the \texttt{pattern} and \texttt{fish} region performs best for both EdgeConnect and LaMa. We also conclude that deep backpropagation does not yield increased results for both EfNet and Inception in these experiments.

Looking at shallow backpropagation fine-tuning, we notice that the worst performance is either just the \texttt{pattern} region (AOT-GAN, DeepFillV2) or just the \texttt{background} region (remaining architectures). Most architectures perform best on just the \texttt{fish} area.

For combinations of areas, we observe that the worst performance is very dependent on the choice of architecture. We also see a split between architectures for the best performing regions, with LaMa, EfNet, and Inception performing best on the combination of the \texttt{fish} and \texttt{pattern} regions. Overall, we see that the classical classification architectures outperform all others, except for LaMa.

From the observations above, we deduce that either the \texttt{pattern} region, which only includes pattern information, or the \texttt{fish} region (which includes parts of the pattern and the perimeter of the fish) is the most important region for most architectures. We also see that only EdgeConnect and LaMa perform better than classical classification architectures (when using shallow backpropagation fine-tuning).

In the end, we propose the usage of the LaMa architecture for the creation of pattern-inspired encoders, since it is the best performing model on just the pattern area, the best performing model on the combination of the fish and pattern areas, and the best performing model on the whole image. It also seems that both deep and shallow fine-tuning of the inpainting pre-trained LaMa encoder are equally effective at extracting embeddings from pattern information included in natural images.

\section{Experiments}
This section describes the implementation of the experiments described in \cref{sec:methodology}. Experiments were run on a machine equipped with 20 Intel Xeon E5-2650v3 cores, 6 NVIDIA GTX 980 Ti (6 GB memory each) GPUs, 2 NVIDIA Titan X (12 GB memory each) GPUs, and 256 GB RAM.

\subsection{Dataset Preparation}

As described in \cref{sec:dataset-collection}, the dataset was taken from the work by Haurum, Karpova, Pedersen \emph{et al}.~\cite{haurum_re-identification_2020}. It includes $2224$ video frames of a well-lit aquarium containing a total of $6$ zebrafish. Each image is annotated with the bounding box coordinates for zebrafish object detection, labels, occlusion, glimmer, and swimming direction.

Zebrafish images were obtained from the video frames by cropping images of unoccluded and non-turning zebrafish to a height and width of $256$ pixels centered on the fish body, and saving them together with the corresponding label.

For the inpainting dataset, $999$ images were randomly selected (stratified over label). For each image, the patterned part of the zebrafish's body was manually masked. Three additional data pairs were generated by applying three image flips (horizontal, vertical, or both), creating a total of four data pairs for each of the original $999$. Labels were subsequently dropped, resulting in a dataset of $3996$ matched images and masks. A random 80:16:4 train:val:test split was performed on the data before training. An example of a paired image and mask can be seen in \cref{fig:inpainting-gt}.

\begin{figure}[tb]
  \centering
  \begin{subfigure}{0.48\linewidth}
    \includegraphics[width=\linewidth]{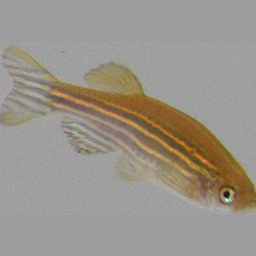}
    \caption{Example of an image in the classification dataset}
    \label{fig:clas-img}
  \end{subfigure}
  \hfill
  \begin{subfigure}{0.48\linewidth}
    \includegraphics[width=\linewidth]{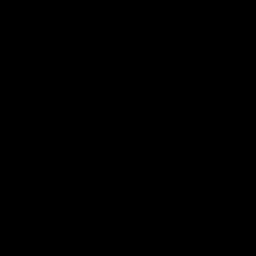}
    \caption{Example of an empty mask in the classification dataset}
    \label{fig:clas-mask}
  \end{subfigure}
  \caption{Example of an image-mask pair in the classification dataset. This pair is annotated with the label $4$.}
  \label{fig:clas-data}
\end{figure}

We selected the remaining $4422$ images for the classification dataset. For inpainting model encoder compatibility, these images were matched to a mask containing only zeros, creating a dataset of $4422$ paired images, masks and labels. A random 80:10:10 train:val:test split was performed on the data before training. An example of an image from the classification dataset and a matched mask can be seen in \cref{fig:clas-data}.

\subsection{Inpainting results}

Inpainting model validation set generator loss (defined as a weighted sum of the loss functions described in \cref{sec:inpainting-method}) was tracked during training and used to determine plateauing, and to regulate early stopping of training. The number of training iterations for the checkpoints used for classification fine-tuning and clustering analysis can be found in \cref{tab:inpainting-checkpoints}.

\begin{table}[tb]
  \caption{Number of training iterations for the inpainting task for each of the four inpainting models. Note that AOT-GAN and LaMa loss is made up of a perceptual, style, adversarial, and regularization (L1) component.}
  \label{tab:inpainting-checkpoints}
  \centering
  \begin{tabular}{@{}llll@{}}
    \toprule
    Model architecture & Lowest validation set generator loss & Batch size & Iterations \\
    \midrule
    AOT-GAN  & 18.673 & 8 & 13000\\
    DeepFillV2 & 0.066796 & 8 & 10000\\
    EdgeConnect & 0.070038 & 8 & 9000\\
    LaMa & 4.2551 & 4 & 12784 \\
  \bottomrule
  \end{tabular}
\end{table}

\begin{figure}[tb]
  \centering
  \begin{subfigure}[t]{0.32\linewidth}
    \includegraphics[width=\linewidth]{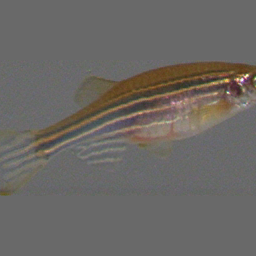}
    \caption{Ground truth image for the inpainting task.}
    \label{fig:gt}
  \end{subfigure}
  \hfill
  \begin{subfigure}[t]{0.32\linewidth}
    \includegraphics[width=\linewidth]{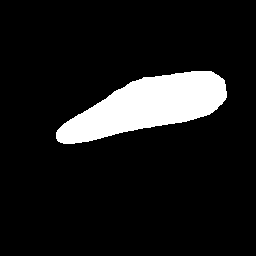}
    \caption{Mask for the inpainting task of the image shown in \cref{fig:gt}}
    \label{fig:mask}
  \end{subfigure}
  \hfill
  \begin{subfigure}[t]{0.32\linewidth}
    \includegraphics[width=\linewidth]{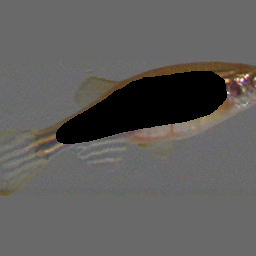}
    \caption{Input for the inpainting task, consisting of the ground truth image (\cref{fig:gt}) multiplied with the inverse of its assigned mask (\cref{fig:mask}).}
    \label{fig:masked}
  \end{subfigure}
  \caption{Ground truth, mask, and input image for one of the images used for the inpainting task}
  \label{fig:inpainting-gt}
\end{figure}

\begin{figure}[tb]
  \centering
  \begin{subfigure}{0.24\linewidth}
    \includegraphics[width=\linewidth]{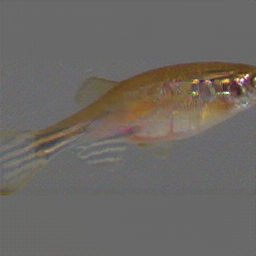}
    \caption{AOT-GAN inpainting result}
    \label{fig:inpainted-aotgan}
  \end{subfigure}
  \hfill
  \begin{subfigure}{0.24\linewidth}
    \includegraphics[width=\linewidth]{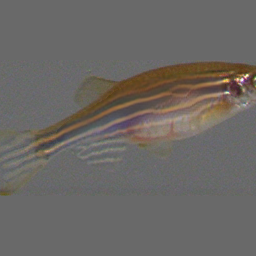}
    \caption{DeepFillV2 inpainting result}
    \label{fig:inpainted-deepfill}
  \end{subfigure}
  \hfill
  \begin{subfigure}{0.24\linewidth}
    \includegraphics[width=\linewidth]{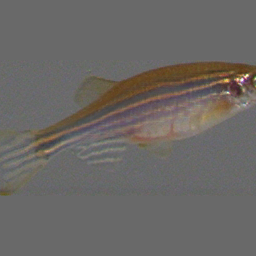}
    \caption{EdgeConnect inpainting result}
    \label{fig:inpainted-edgeconnect}
  \end{subfigure}
  \hfill
  \begin{subfigure}{0.24\linewidth}
    \includegraphics[width=\linewidth]{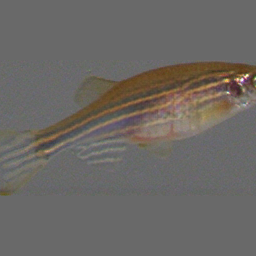}
    \caption{LaMa inpainting result}
    \label{fig:inpainted-lama}
  \end{subfigure}
  \caption{Output of inpainting using all four selected inpainting models}
  \label{fig:qual-inpainting}
\end{figure}

Using the checkpoints defined in \cref{tab:inpainting-checkpoints}, a qualitative visual analysis was performed on the inpainting results. The ground truth image can be seen in \cref{fig:inpainting-gt}. Inpainting results for AOT-GAN, DeepFillV2, EdgeConnect, and LaMa can be seen in \cref{fig:qual-inpainting}.

We observe that the inpainted masks visually resemble the ground-truth pattern for all inpainting models but AOT-GAN, which removes the pattern from the fish. Visually, LaMa produces the best pattern reconstruction. Given the inpainting results, we assume that DeepFillV2, EdgeConnect, and LaMa successfully extracted pattern information from the zebrafish.

\begin{figure}[tb]
  \centering
  \begin{subfigure}{0.49\linewidth}
    \includegraphics[width=\linewidth]{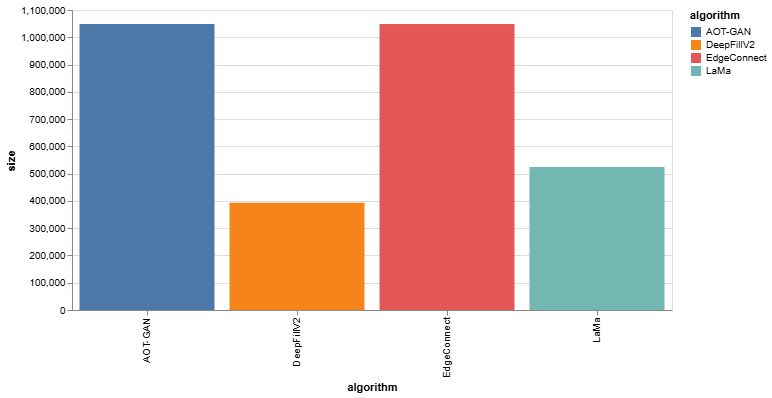}
    \caption{Number of embedding features per inpainting algorithm}
    \label{fig:embedding-size}
  \end{subfigure}
  \hfill
  \begin{subfigure}{0.49\linewidth}
    \includegraphics[width=\linewidth]{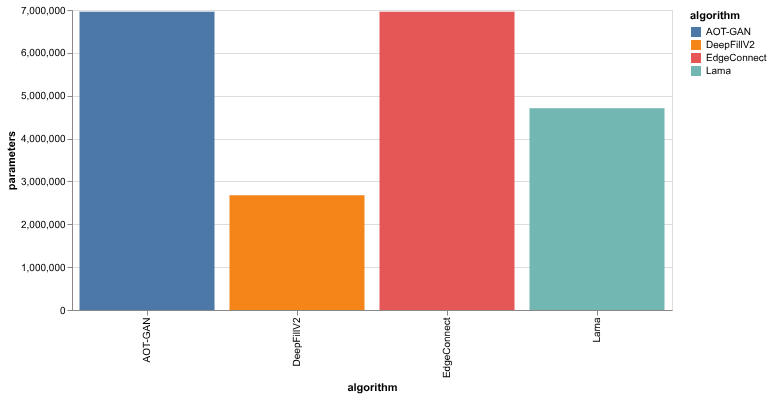}
    \caption{Number of encoder parameters per isolated encoder}
    \label{fig:parameter-size}
  \end{subfigure}
  \caption{Differences in number of embedding features and encoder parameters between AOT-GAN, DeepFillV2, EdgeConnect and LaMa.}
  \label{fig:encoder-embedding-size}
\end{figure}

After inpainting training, we generated embeddings for every image in the classification test set. We observed a strong variation in the embeddings' dimensions, since encoder embedding spaces vary in number of features. Consequently, inpainting architecture encoders also vary greatly in number of parameters. A comparison between the different models can be seen in \cref{fig:embedding-size}.

\subsection{Classification results}

\begin{figure}[tb]
  \centering
  \begin{subfigure}{0.32\linewidth}
    \includegraphics[width=\linewidth]{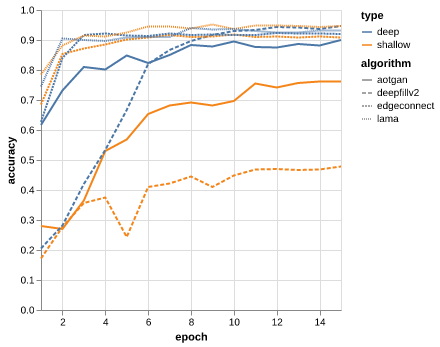}
    \caption{Accuracy computed on the validation set during classification training}
    \label{fig:val-acc}
  \end{subfigure}
  \hfill
  \begin{subfigure}{0.32\linewidth}
    \includegraphics[width=\linewidth]{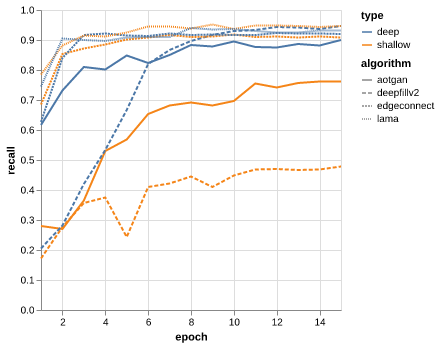}
    \caption{Recall computed on the validation set during classification training}
    \label{fig:val-rec}
  \end{subfigure}
  \begin{subfigure}{0.32\linewidth}
    \includegraphics[width=\linewidth]{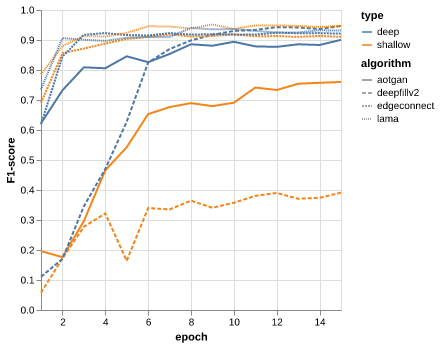}
    \caption{F1-score computed on the validation set during classification training}
    \label{fig:val-f1}
  \end{subfigure}
  \begin{subfigure}{0.32\linewidth}
    \includegraphics[width=\linewidth]{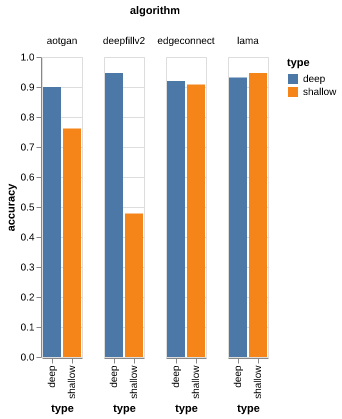}
    \caption{Accuracy computed on the test set after classification training}
    \label{fig:test-acc}
  \end{subfigure}
  \hfill
  \begin{subfigure}{0.32\linewidth}
    \includegraphics[width=\linewidth]{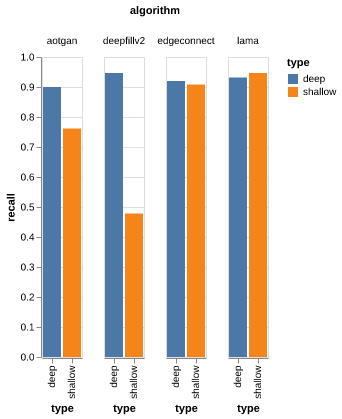}
    \caption{Recall computed on the test set after classification training}
    \label{fig:test-rec}
  \end{subfigure}
  \begin{subfigure}{0.32\linewidth}
    \includegraphics[width=\linewidth]{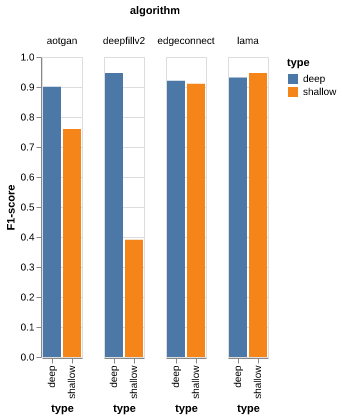}
    \caption{F1-score computed on the test set after classification training}
    \label{fig:test-f1}
  \end{subfigure}
  \caption{Classification quality metrics using the inpainting pre-trained encoder and a classification head. Shallow refers to shallow backpropagation, and deep refers to deep backpropagation.}
  \label{fig:classification-plots}
\end{figure}

Classification model accuracy, recall, and F1-score were tracked on the validation set during training, and computed on the test set after training. Results can be seen in \cref{fig:classification-plots}.

When comparing training performance between shallow and deep backpropagation during classification fine-tuning, we observe a faster increase in classification accuracy for all encoder backbones when using deep backpropagation, except for the LaMa encoder. We also observe that the deep backpropagation encoder outperforms the shallow backpropagation encoder on the test set for all models except LaMa.

\begin{figure}[tb]
  \centering
  \begin{subfigure}{0.32\linewidth}
    \includegraphics[width=\linewidth]{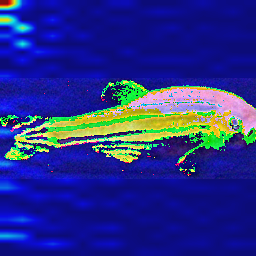}
    \caption{GradCAM overlay of individual 1 after 15 epochs}
    \label{fig:lama-gradcam-shallow-1}
  \end{subfigure}
  \hfill
  \begin{subfigure}{0.32\linewidth}
    \includegraphics[width=\linewidth]{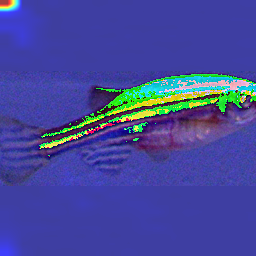}
    \caption{GradCAM overlay of individual 2 after 15 epochs}
    \label{fig:lama-gradcam-shallow-2}
  \end{subfigure}
  \begin{subfigure}{0.32\linewidth}
    \includegraphics[width=\linewidth]{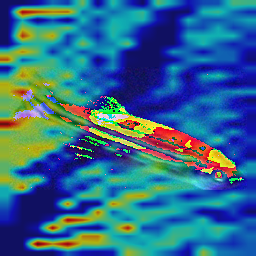}
    \caption{GradCAM overlay of individual 3 after 15 epochs}
    \label{fig:lama-gradcam-shallow-3}
  \end{subfigure}
  \begin{subfigure}{0.32\linewidth}
    \includegraphics[width=\linewidth]{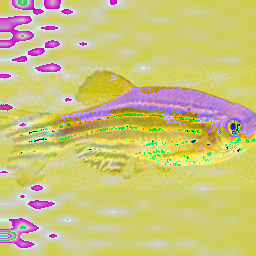}
    \caption{GradCAM overlay of individual 4 after 15 epochs}
    \label{fig:lama-gradcam-shallow-4}
  \end{subfigure}
  \hfill
  \begin{subfigure}{0.32\linewidth}
    \includegraphics[width=\linewidth]{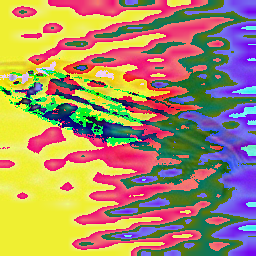}
    \caption{GradCAM overlay of individual 5 after 15 epochs}
    \label{fig:lama-gradcam-shallow-5}
  \end{subfigure}
  \begin{subfigure}{0.32\linewidth}
    \includegraphics[width=\linewidth]{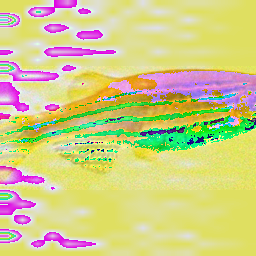}
    \caption{GradCAM overlay of individual 6 after 15 epochs}
    \label{fig:lama-gradcam-shallow-6}
  \end{subfigure}
  \caption{GradCAM-overlaid zebrafish image results after 15 epochs of shallow backpropagation classification fine-tuning for each individual zebrafish in the dataset}
  \label{fig:lama-gradcam-shallow}
\end{figure}

\begin{figure}[tb]
  \centering
  \begin{subfigure}{0.32\linewidth}
    \includegraphics[width=\linewidth]{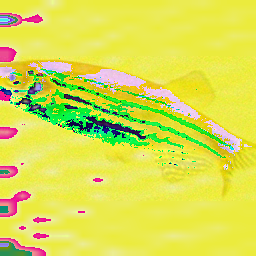}
    \caption{GradCAM overlay of individual 6 before classification fine-tuning}
    \label{fig:lama-gradcam-deep-0}
  \end{subfigure}
  \hfill
  \begin{subfigure}{0.32\linewidth}
    \includegraphics[width=\linewidth]{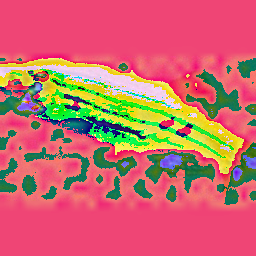}
    \caption{GradCAM overlay of individual 6 after 3 epochs}
    \label{fig:lama-gradcam-deep-3}
  \end{subfigure}
  \begin{subfigure}{0.32\linewidth}
    \includegraphics[width=\linewidth]{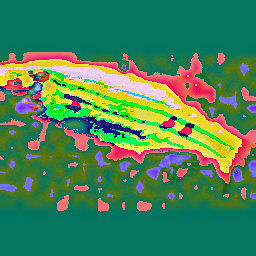}
    \caption{GradCAM overlay of individual 6 after 6 epochs}
    \label{fig:lama-gradcam-deep-6}
  \end{subfigure}
  \begin{subfigure}{0.32\linewidth}
    \includegraphics[width=\linewidth]{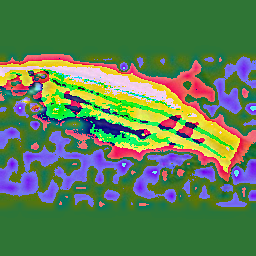}
    \caption{GradCAM overlay of individual 6 after 9 epochs}
    \label{fig:lama-gradcam-deep-9}
  \end{subfigure}
  \hfill
  \begin{subfigure}{0.32\linewidth}
    \includegraphics[width=\linewidth]{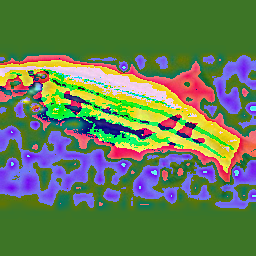}
    \caption{GradCAM overlay of individual 6 after 12 epochs}
    \label{fig:lama-gradcam-deep-12}
  \end{subfigure}
  \begin{subfigure}{0.32\linewidth}
    \includegraphics[width=\linewidth]{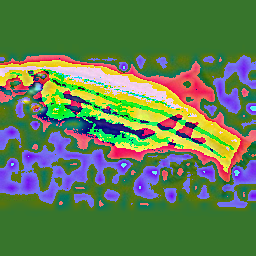}
    \caption{GradCAM overlay of individual 6 after 15 epochs}
    \label{fig:lama-gradcam-deep-15}
  \end{subfigure}
  \caption{GradCAM-overlaid zebrafish image results during deep backpropagation classification fine-tuning for zebrafish individual 6}
  \label{fig:lama-gradcam-deep}
\end{figure}

GradCAM backpropagation overlays of node activations were also plotted for both shallow (\cref{fig:lama-gradcam-shallow}) and deep backpropagation (\cref{fig:lama-gradcam-deep}). We present LaMa GradCAM overlays here.

After classification fine-tuning, embeddings were generated for every image in the classification test set. These embeddings are further analysed in \cref{sec:embedding-analysis}.

\subsection{Embedding clustering results} \label{sec:embedding-analysis}

\begin{figure}[tb]
  \centering
  \begin{subfigure}{0.48\linewidth}
    \includegraphics[width=\linewidth]{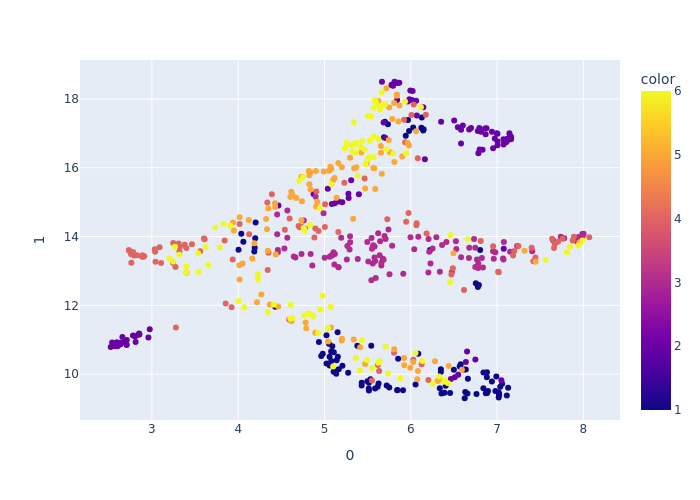}
    \caption{UMAP visualization of LaMa embeddings after inpainting training}
    \label{fig:lama-raw-umap}
  \end{subfigure}
  \hfill
  \begin{subfigure}{0.48\linewidth}
    \includegraphics[width=\linewidth]{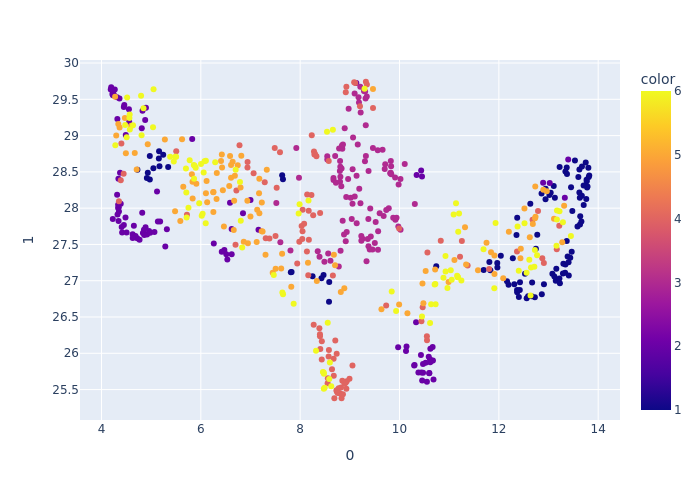}
    \caption{UMAP visualization of LaMa embeddings after classification fine-tuning}
    \label{fig:lama-deep-umap}
  \end{subfigure}
  \hfill
  \caption{UMAP visualizations of the embedding space before (\cref{fig:lama-raw-umap}) and after (\cref{fig:lama-deep-umap}) classification fine-tuning.}
  \label{fig:lama-umap}
\end{figure}

Generated embeddings were first visually inspected through the dimensionality reduction techniques PCA~\cite{mackiewicz_principal_1993}, T-SNE~\cite{maaten_visualizing_2008} and UMAP~\cite{mcinnes_umap_2020}. We present LaMa UMAP visualizations in \cref{fig:lama-umap}.

Afterwards, embeddings were clustered through the k-means algorithm with $n=6$, and k-means++ initialization. After clustering, the Silhouette Score, Davies-Bouldin Index, Calinski-Harabasz Index, Adjusted Rand Index, and Mutual Information were computed. These can be seen in \cref{tab:clustering-metrics}.

\begin{table}
  \label{tab:clustering-metrics}
  \centering
  \caption{Clustering metrics for encoders frozen after inpainting, and encoders refined through classification fine-tuning (ref). Rounded to three decimals. Best algorithm is denoted in \textbf{bold} for each metric.}
  \begin{tabular}{llllll}
    \toprule
    Encoder & AdjRand & MutInfo & Silhouette & daviesBouldin & calinskiHarabasz \\
    \midrule
    aotgan & 0.127 & 0.339 & 0.060 & 3.224 & 36.558 \\
    ref-aotgan & \textbf{0.228} & \textbf{0.572} & \textbf{0.181} & \textbf{1.945} & \textbf{84.635} \\
    deepfillv2 & 0.114 & 0.315 & 0.033 & 4.154 & 22.690 \\
    ref-deepfillv2 & 0.140 & 0.322 & 0.033 & 3.696 & 27.429 \\
    edgeconnect & 0.137 & 0.324 & 0.049 & 3.469 & 23.148 \\
    ref-edgeconnect & 0.157 & 0.418 & 0.038 & 3.792 & 19.092 \\
    lama & 0.136 & 0.331 & 0.042 & 3.425 & 27.705 \\
    ref-lama & 0.117 & 0.284 & 0.031 & 4.051 & 19.171 \\
    \bottomrule
  \end{tabular}
\end{table}

We observe that the classification fine-tuned (deep backpropagation) AOT-GAN encoder outperforms all other encoders in terms of clustering capability for all shown metrics.

\section{Conclusion}

In this paper, we explored the effect of inpainting as an auxiliary task to enhance individual identification using deep learning based on animal skin patterns.

We've shown that inpainting pre-training leads to the successful extraction of informative features from images, which can subsequently be used for individual classification with only minimal fine-tuning. Our ablation study shows that the most important image region for feature extraction depends heavily on the chosen inpainting model during the pre-training task. We've also shown that the combination of a LaMa pre-trained encoder and a fine-tuned linear classification head outperforms previous models on this classification task, regardless of the used backpropagation method or ablated regions.

Clustering analysis of the inpainting models' embedding spaces revealed refined AOT-GAN to be the best performing model for all computed metrics. This shows a disconnect between quality of inpainting, classification performance, and embedding clustering quality, since AOT-GAN did not outperform other models during both inpainting and classification.

Given the results in this paper, potential future work on the creation of pattern-based embeddings should focus on improving the amount of information embedded from the patterned region of the input image. This could be done either through tweaking the pre-training task, for example by changing the types of masks used during inpainting or enlarging the dataset, or through the creation of a novel learning paradigm or model architecture for creating pattern-inspired embeddings.

\clearpage
\bibliographystyle{unsrt}
\bibliography{bibliography/main}
\end{document}